\pdfoutput=1
\documentclass[12pt]{l4dc2027} 

\usepackage{times}
\usepackage{mathtools}
\usepackage{booktabs}
\usepackage{placeins}

\newcommand{\sign}{\operatorname{sign}}
\newcommand{\E}{\mathbb{E}}

\title[Learning to Tune with Biased Gradients]{Learning to Tune with Biased Gradients:
Iterative Feedback Tuning of Model-Free Control on Throttle Valves}

\coltauthor{%
 \Name{F\'elix Mart\'inez-Jim\'enez} \Email{fdmtz01@outlook.com}\\
 \Name{Sander Miller-Murphy} \Email{sander.miller.murphy@gmail.com}\\
 \Name{Tobie Ondoa-Essomba} \Email{ondoadimitri3@gmail.com}\\
 \Name{Emmanuel Witrant} \Email{emmanuel.witrant@univ-grenoble-alpes.fr}\\
 \addr Univ.\ Grenoble Alpes, CNRS, GIPSA-lab, Grenoble, France}

\begin{document}
\maketitle

\begin{abstract}
Iterative feedback tuning (IFT) learns the parameters of a fixed-structure controller by stochastic gradient descent. At each iteration, two closed-loop experiments are run on the plant: the first one measures the tracking error, and the second one, driven by this error, provides the sensitivities from which the gradient of the cost is estimated. This estimate is unbiased for linear plants only. We study the biased case on automotive electronic throttle valves, whose friction and stiction vary between valves and runs, controlled by an intelligent proportional (model-free) controller, which is an incremental PI controller. We show that the mismatch of the controller state at the start of the two experiments, which stiction makes unavoidable, adds an error inversely proportional to the amplitude of the second experiment. On digital twins of ten valves, the IFT gradient is a descent direction in only 69\% of the low-cost region, friction being the cause, and all update rules plateau and then drift. On two hardware valves, the best update rule depends on the noise structure of the gradient. On the first valve, where one gradient component is buried in noise, Adam outperforms Gauss--Newton, its normalisation and its momentum each improving one parameter. On the second valve, where both components are consistent, the same normalisation amplifies a direction that Gauss--Newton rightly discounts, and Adam drifts to a controller worse than the initial one. The twins reproduce the mechanism but not this ranking.
\end{abstract}


\section{Introduction}

Data-driven controller tuning casts the design of a fixed-structure controller as the minimisation of a closed-loop cost $J(\vartheta)$ that can only be queried through experiments on the plant. Iterative feedback tuning (IFT) \citep{hjalmarsson1998iterative,hjalmarsson2002overview} is the stochastic-gradient member of this family: closed-loop experiments provide at each iteration an estimate $\hat g$ of $\nabla J$, which is unbiased when the plant is linear time-invariant and the noise is zero-mean. Other approaches avoid gradients, either by solving a model-reference identification problem from a single data set, as virtual reference feedback tuning \citep{campi2002virtual} and correlation-based tuning \citep{karimi2008model}, or by treating $J$ as a black box, as Bayesian optimisation \citep{berkenkamp2016safe}. IFT remains attractive because it needs no model, scales to many parameters and has the structure of the stochastic optimisation methods of machine learning, including adaptive ones such as Adam \citep{Kingma2014AdamAM}. IFT has been applied to the intelligent PID controllers of model-free control in simulation \citep{baciu2021iterative} and with Adam-like updates \citep{baciu2023iterative}. \citet{munteanu2025practical} report practical difficulties of its real-time implementation, in particular its sensitivity to noisy gradient estimates and to the step size.

The unbiasedness of $\hat g$, however, rests on two assumptions that physical plants routinely violate: the closed-loop map must be linear, and the experiments of one iteration must start from the same state. On nonlinear plants, the classical recommendation is a small second experiment that probes the linearisation along the first one \citep{sjoberg1996model,hjalmarsson1998nonlinear}. Mechanical systems with Coulomb friction and stiction have no such linearisation, and their equilibrium is a set rather than a point. For a generic biased gradient oracle, \citet{ajalloeian2020convergence} show that stochastic gradient descent only converges to a neighbourhood of a stationary point whose size scales with the bias. What is missing is an account, for a concrete plant, of where the bias comes from, how large it is where tuning actually operates, how it interacts with the update rule, and whether a digital twin can predict it.

We address these questions on electronic throttle valves (ETVs), low-cost servos with a preloaded return spring, Coulomb friction and stiction. Their friction is strongly asymmetric, which motivated dedicated models \citep{Pujol2016} and adaptive or friction-compensating controllers \citep{Canudas2001,Jiao2014}, and it varies significantly between valves of the same reference and between repetitions of the same test \citep{witrant2023data}. The controller is the intelligent proportional (iP) controller of model-free control \citep{fliess2013model}, whose two gains are usually tuned by trial and error \citep{yahagi2022non} and which has been applied to the same benchmark by \citet{belhadjoudja2023intelligent}. Preliminary real-time experiments with IFT and an Adam-based variant on this benchmark improved a hand-tuned controller but stalled after a few iterations and showed no benefit of Adam; this paper explains both observations.

Section~\ref{sec:ift} shows that the initial-state mismatch caused by stiction adds to the IFT gradient an error proportional to $1/c$, where $c$ is the amplitude of the second experiment. Section~\ref{sec:twin} quantifies the bias on digital twins of ten valves identified from previous data \citep{witrant2023data} and isolates its cause, and Section~\ref{sec:opt} compares Gauss--Newton (GN), trust-region, sign and Adam updates on these twins. Section~\ref{sec:bench} reports hardware experiments on two valves: repeatability, a finite-difference check of the gradient, repeated tuning runs that separate the effects of normalisation and momentum, and a comparison with the twin under the same protocol. The firmware, the host code, the data and the scripts reproducing all figures and tables are publicly available \citep{martinez2026code}.

\section{Model-free control and iterative feedback tuning}\label{sec:ift}

Let $y$ be the measured valve angle, $r$ its reference and $e=y-r$ the tracking error. Following \citet{belhadjoudja2023intelligent}, the ultra-local model of the iP controller is $\dot e=F+\alpha u$, where $F$ gathers everything unknown and $\alpha>0$ is a design parameter. With the backward-Euler estimate $\hat F_k=(e_k-e_{k-1})/T_s-\alpha u_{k-1}$ and the control law $u_k=-(Ke_k+\hat F_k)/\alpha$, where $K>0$ sets the desired error dynamics $\dot e=-Ke$, the controller reads
\begin{equation}
u_k=u_{k-1}-q_0e_k-q_1e_{k-1},\qquad
q_0=\frac{1+T_sK}{\alpha T_s},\quad q_1=-\frac{1}{\alpha T_s}.
\label{eq:ip}
\end{equation}
It is therefore an incremental PI controller, $u=C(z)(r-y)$ with $C(z)=\dfrac{q_0+q_1z^{-1}}{1-z^{-1}}=K_p+\dfrac{K_iT_s}{1-z^{-1}}$, where $K_p=-q_1=1/(\alpha T_s)$ and $K_i=(q_0+q_1)/T_s=K/(\alpha T_s)$ are the proportional and integral gains. The map is invertible, $K=-(q_0+q_1)/(q_1T_s)$ and $\alpha=-1/(q_1T_s)$, and $K,\alpha>0$ if and only if $q_1<0<q_0+q_1$. With $T_s=50$\,ms, the hand-tuned controller of the bench, $(K,\alpha)=(5,2.5)$, corresponds to $(q_0,q_1)=(10,-8)$.

Because of the spring preload, the IFT experiments are run in deviation around an operating angle $\theta_A$, from an equilibrium $(\theta_A,u_A)$ reached in closed loop. Experiment~1 applies a reference $r_1$ consisting of one up and one down step, $\theta_A\to\theta_B\to\theta_A$ ($r_1=\theta_B$ on samples 10 to 109 of $N=200$, i.e.\ 5\,s out of 10\,s), and records the output $y_1$, the error $e_1=y_1-r_1$ and the input deviation $\Delta u_1=u_1-u_A$. The cost and its gradient with respect to $\rho=[q_0\ q_1]^\top$ are, for $N$ samples,
\begin{equation}
J(\rho)=\frac{1}{2N}\sum_{k=1}^N\bigl(e_{1,k}^2+\lambda\Delta u_{1,k}^2\bigr),\qquad
\nabla J=\frac1N\sum_{k=1}^N\bigl(e_{1,k}\,\psi^y_k+\lambda\,\Delta u_{1,k}\,\psi^u_k\bigr),
\label{eq:dJ}
\end{equation}
with $\lambda=0.01$ and the output and input sensitivities $\psi^y=\partial y/\partial\rho$ and $\psi^u=\partial u/\partial\rho$. For a linear plant modelled by the transfer function $G(z)$, with the sensitivity function $S=(1+GC)^{-1}$ and the complementary sensitivity function $T=GCS$, these sensitivities are $\psi^y=-\Phi\,T e_1$ and $\psi^u=-\Phi\,CS e_1$, where $\Phi(z)=C^{-1}\partial C/\partial\rho=(q_0+q_1z^{-1})^{-1}[1\ \ z^{-1}]^\top$ \citep{hjalmarsson1998iterative}. They involve the unknown $T$ and $CS$ applied to a signal, which is exactly what a closed-loop experiment computes. To obtain them without a model, experiment~2 therefore feeds the error of experiment~1 back as reference, $r_2=\theta_2-c\,e_1$, where $c>0$ is its excitation amplitude ($c=1$ and $\theta_2=\theta_A$ in standard IFT), and the sensitivities are estimated by
\begin{equation}
\hat\psi^y=\frac1c\,\Phi\,(y_2-\theta_2),\qquad \hat\psi^u=\frac1c\,\Phi\,(u_2-u_{2,0}),
\label{eq:psi}
\end{equation}
where $u_{2,0}$ is the input at the start of experiment~2. The gradient estimate and the Gauss--Newton approximation of the Hessian of $J$ are then
\begin{equation}
\hat g=\frac1N\sum_{k=1}^N\bigl(e_{1,k}\,\hat\psi^y_k+\lambda\,\Delta u_{1,k}\,\hat\psi^u_k\bigr),\qquad
\hat R=\frac1N\sum_{k=1}^N\bigl(\hat\psi^y_k\hat\psi^{y\top}_k+\lambda\,\hat\psi^u_k\hat\psi^{u\top}_k\bigr),
\label{eq:gR}
\end{equation}
and the standard IFT update is the Gauss--Newton step $\rho_{i+1}=\rho_i-\gamma\hat R^{-1}\hat g$. The pole of $\Phi$ is $1/(1+T_sK)$, so that the filters are stable if and only if $K>0$. Since only deviations are used, $\theta_2$ may differ from $\theta_A$, a special case of the filtered second experiment of \citet{prochazka2005iterative}.

IFT assumes that both experiments start from the same state, which a plant with stiction does not allow. The valve rests at the target angle for any net torque within the stiction band, so that the integral state of the controller at the start of experiment~2, $u_{2,0}$, differs from $u_A$ by an amount $\delta$ that depends on the history and is not controlled. Seen from a linear model, the valve is then held by a friction torque equivalent to an input offset $d=-\delta$, which is released as soon as the valve moves. The following result quantifies the effect of such an offset; $H$ denotes the unit step (Heaviside) sequence.

\begin{proposition}\label{prop:bias}
Let the plant be $y-\theta_2=G(z)(u-u_{2,0}+d\,H)$ during experiment~2, where $G$ is linear and $d$ is a constant offset, the loop being stable. Then the estimates \eqref{eq:psi} satisfy $\hat\psi^y=\psi^y+\frac dc\,\Phi\,GS\,H$ and $\hat\psi^u=\psi^u-\frac dc\,\Phi\,T\,H$, and
\begin{equation}
\hat g-\nabla J=\frac dc\,b,\qquad b=\frac1N\sum_{k=1}^N\Bigl(e_{1,k}\,(\Phi\,GS\,H)_k-\lambda\,\Delta u_{1,k}\,(\Phi\,T\,H)_k\Bigr).
\label{eq:bias}
\end{equation}
\end{proposition}

The proof, given in Appendix~\ref{app:proof}, follows from linearity. The error scales as $1/c$, so that the small second experiment recommended for smooth nonlinear plants amplifies it, whereas a large one drives the valve into the input saturation and the stops. The vector $b$ depends only on $\rho$, $G$ and experiment~1: if $\E[d]\neq0$, for instance when the valve tends to stick on the same side, the error is a systematic bias, and if $\E[d]=0$ it is an additional variance that does not decrease with the length of the experiment. Finally, the offset term is a step response, concentrated at low frequencies where the gain of $\Phi$ is $1/(q_0+q_1)=\alpha/K$, the inverse of the integral gain; the error is thus amplified precisely for the sluggish controllers that a tuning algorithm explores when it reduces $K$ or increases $\alpha$. In reality the offset is not constant, since the valve may stick again during the experiment, which adds a state-dependent error; Section~\ref{sec:twin} quantifies the total effect.

The parametrisation matters for coordinate-wise methods. The first update of Adam moves every coordinate by approximately its step size $\mu$, whatever the magnitude of the gradient. In $(q_0,q_1)$ coordinates, the integral gain $K_iT_s=q_0+q_1=K/\alpha$ is the small difference of two large numbers, $10-8=2$ at the hand-tuned point, so that a first step with $\mu=1$ that decreases $q_0$ and $q_1$ by one unit each cancels the integral action. We therefore optimise $\vartheta=(\log K,\log\alpha)$, which keeps $K$ and $\alpha$ positive and makes the steps relative, with $\hat g_\vartheta=Q^\top\hat g$, $\hat R_\vartheta=Q^\top\hat RQ$ and $Q=\partial\rho/\partial\vartheta$. Five update rules are compared. GN, $\vartheta_{i+1}=\vartheta_i-\gamma\hat R_\vartheta^{-1}\hat g_\vartheta$, is the standard IFT update, and clipped GN limits its step to $\|\vartheta_{i+1}-\vartheta_i\|\le0.2$. The sign step $\vartheta_{i+1}=\vartheta_i-h\sign(\hat g_\vartheta)$ \citep{bernstein2018signsgd} normalises each coordinate but has no memory. Adam \citep{Kingma2014AdamAM} combines this normalisation with momentum,
\begin{equation}
m_i=\beta_1m_{i-1}+(1-\beta_1)\hat g_{\vartheta,i},\quad
v_i=\beta_2v_{i-1}+(1-\beta_2)\hat g_{\vartheta,i}^{\,2},\quad
\vartheta_{i+1}=\vartheta_i-\mu\,\frac{m_i/(1-\beta_1^i)}{\sqrt{v_i/(1-\beta_2^i)}+\epsilon},
\label{eq:adam}
\end{equation}
where squares, square roots and ratios are taken element-wise, $m_0=v_0=0$, $\beta_1=0.9$, $\beta_2=0.999$ and $\epsilon=10^{-8}$. ``Adam-GN'' applies \eqref{eq:adam} to the GN direction $\hat R_\vartheta^{-1}\hat g_\vartheta$ instead of $\hat g_\vartheta$, as in the preliminary experiments. Comparing the sign step and Adam separates the effects of normalisation and momentum.

\section{Digital twins and reliability of the gradient}\label{sec:twin}

The benchmark of \citet{witrant2023data} consists of throttle valves of the same reference whose return spring opens the valve; a DC motor driven by a PWM signal in $[0,255]$ closes it, and the angle $\theta$ is measured by a magnetic sensor at $T_s=50$\,ms. Each valve has a PRBS record and a slow PWM staircase with several up--down cycles. Following \citet{Pujol2016}, the shaft dynamics are
\begin{equation}
\dot\theta=\omega,\qquad
\dot\omega=-a_1(\theta-\theta_s)-a_2^{\sigma}\omega-a_3^{\sigma}\sign(\omega)+a_4 v,
\label{eq:twin}
\end{equation}
with $v=-\mathrm{PWM}$ and $\sigma\in\{+,-\}$ the direction of motion. We extended the initial model of \citet{Pujol2016} with a spring equilibrium $\theta_s$ beyond the upper stop (preload), a Karnopp stiction model \citep{karnopp1985} with direction-dependent breakaway levels $a_{3s}^\pm$, and inelastic stops. The eleven parameters are fitted by differential evolution followed by the Nelder--Mead (downhill simplex) method on the PRBS and the first staircase cycle, the other cycles being kept for cross-validation (Appendix~\ref{app:twin}). The simulation RMS errors are 0.9--3.5$^\circ$ on the PRBS and 4.9--13.9$^\circ$ in cross-validation. The latter are dominated by the positions at which the valve sticks, which vary from cycle to cycle in the data themselves: the closed position of one valve varies between 2$^\circ$ and 30$^\circ$. The twins are therefore behavioural models of the stick--slip phenomenology.

\begin{figure}[!t]
\centering
\includegraphics[width=\textwidth]{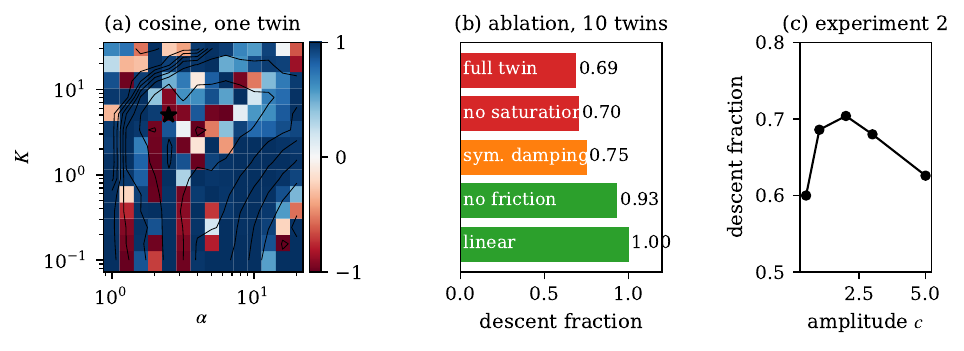}
\caption{Twins. (a) Cosine between the IFT and true gradients for one valve ($\star$: hand-tuned controller; contours: $\log J$). (b) Descent fraction averaged over the ten twins when removing each nonlinearity. (c) Effect of the amplitude $c$ of experiment~2.}
\label{fig:grad}
\end{figure}

On each twin, the IFT gradient $\hat g(\vartheta)$ ($c=1$, experiment 30--60--30$^\circ$) is compared with the gradient $\nabla J(\vartheta)$ of the true cost, obtained by central finite differences, on a $10\times10$ log-spaced grid $\mathcal G$ covering $K\in[0.1,30]$ and $\alpha\in[1,20]$. The low-cost half of the grid is $\mathcal G_-=\{\vartheta\in\mathcal G: J(\vartheta)\le\operatorname{median}_{\mathcal G}J\}$, and we define the \emph{descent fraction}
\begin{equation}
D=\frac{1}{|\mathcal G_-|}\sum_{\vartheta\in\mathcal G_-}\mathbf 1\bigl[\hat g(\vartheta)^\top\nabla J(\vartheta)>0\bigr],
\end{equation}
i.e.\ the proportion of low-cost points where an infinitesimal step along $-\hat g$ decreases $J$. On the linearised twins, the cosine between both gradients exceeds $0.99999$ at all 177 points where the loop is stable, which validates the implementation. On the actual twins, the descent fraction is only 0.69 on average (0.54--0.80 across valves), and the wrong directions concentrate in the valley of the cost around the hand-tuned controller (Fig.~\ref{fig:grad}a), i.e.\ where a tuning algorithm operates.

Removing one nonlinearity at a time identifies the cause (Fig.~\ref{fig:grad}b). The input saturation plays no role (0.70) and symmetrising the damping helps slightly (0.75), whereas removing Coulomb friction and stiction restores a descent fraction of 0.93, and of 1.00 for nine valves out of ten. Consistently with Proposition~\ref{prop:bias}, a smaller second experiment is counter-productive (Fig.~\ref{fig:grad}c): the descent fraction falls to 0.60 for $c=0.5$, where the valve remains stuck more often, while $c=5$ drives it into the saturation and the stops (0.63). No amplitude removes the bias.

\section{Update rules under biased gradients}\label{sec:opt}

Each rule is run for 40 iterations from $(K,\alpha)=(5,2.5)$ on the ten twins in three scenarios: a nominal one; a ``noise'' scenario, with a measurement noise of standard deviation 0.5$^\circ$ and friction coefficients scaled by $e^{0.2\xi}$, $\xi\sim\mathcal N(0,1)$, drawn independently for each experiment; and a ``harsh'' scenario with 2$^\circ$ and $e^{0.4\xi}$. Eight seeds are used per valve. Performance is measured on the noise-free nominal twin by the normalised gap $g_i=(J(\vartheta_i)-J^\star)/(J(\vartheta_0)-J^\star)$, $J^\star$ being the minimum over a $31\times31$ grid, and a run is counted as failed if the parameters become inadmissible or the validation error explodes.

For all well-tuned rules, the median gap reaches about 0.5 within 10--15 iterations and then stalls (Table~\ref{tab:opt} and Fig.~\ref{fig:opt}, Appendix~\ref{app:opt}), which is the behaviour expected from biased stochastic gradient descent \citep{ajalloeian2020convergence}. Continuing the iterations is harmful: after 40 iterations, 19--46\% of the runs end with a higher cost than the initial controller, even without noise. Rejecting the steps that increase the measured cost removes this tail but also stops the progress, since the measured cost is itself corrupted by the friction variability. Large GN steps ($\gamma\ge0.3$) fail in 20--51\% of the runs; clipping removes the failures but not the drift.

With $\mu=0.02$, Adam never fails, reaches the same median gap as GN and has a much smaller tail, the 90th percentile of $g_{40}$ being 1.76 against 3.69 for GN ($\gamma=0.1$) under noise, and 2.29 against 6.90 in the harsh scenario. Under moderate noise, Adam is significantly better than GN with $\gamma=0.1$ but not than the sign step, so that its benefit comes from the bounded, normalised step. In the harsh scenario, it also outperforms the sign step, because the first moment averages out when the biased gradient changes sign around the valley, which shrinks the updates and delays the drift (Appendix~\ref{app:opt}). Adam-GN brings no benefit, and far from the optimum the fixed relative step of Adam is a drawback.

Finally, decreasing $J$ did not improve a validation test spanning 15--65$^\circ$ (median RMS ratio 0.94--1.16), because the IFT experiment (30--60$^\circ$) excites a different friction regime: the IFT experiment must span the operating range of interest, which we do on the bench.

\section{Hardware experiments}\label{sec:bench}

\subsection{Experiment server and protocol}

The firmware turns the Arduino of the benchmark into an experiment server driven by a Python host over the serial link. It implements the iP law \eqref{eq:ip} with the saturated $u_{k-1}$ in $\hat F_k$, which provides anti-windup, and fixed-rate scheduling. Each run starts with a 3\,s closed-loop settling phase at its initial angle, and experiment~2 is generated on board from the stored $e_1$, so that both experiments share the same sampling instants. The host computes the estimates and the updates and, every second iteration, runs the validation test of \citet{belhadjoudja2023intelligent}, a 15/65$^\circ$ square reference of 20\,s. The experiments were run on valves 9 and 0 of the benchmark, called A and B below, starting from the hand-tuned controller; a PI controller previously used on this bench \citep{witrant2023data}, equivalent to $(K,\alpha)=(4.41,5.08)$, serves as a second reference.

\subsection{Repeatability: the cost is set by where the valve sticks}

We measure the dispersion of the cost over $n$ repetitions with fixed controller gains by its coefficient of variation $\mathrm{CV}=s_J/\bar J$, where $\bar J$ and $s_J$ are the sample mean and standard deviation of the measured costs. With the 30--60$^\circ$ experiment of the twin study, ten repetitions at the hand-tuned controller give $\mathrm{CV}=19\%$ on valve~A. This dispersion is not due to measurement noise: $J$ is determined by the controller state $u_A$ at the end of the settling phase (correlation $-0.90$), which spreads over the stiction band with a standard deviation of 17 PWM units for the same angle. The two experiments of one iteration therefore start from integral states that differ by 17.5 PWM units on average, so that the offset $d$ of Proposition~\ref{prop:bias} is observed directly. Running experiment~1 over the validation range (15--65$^\circ$) and experiment~2 around $\theta_2=40^\circ$, away from both stops, reduces the CV to 6.3\%. The price is a less informative gradient. For each of the ten repetitions, we compute the angle between its estimate $\hat g$ and the mean of the ten estimates: $0^\circ$ means that the estimate points in the mean direction, $90^\circ$ that a step along it does not move the parameters in the mean direction at all. At 30--60$^\circ$, the angles are small (median $7^\circ$, maximum $33^\circ$), so that every iteration moves the parameters in nearly the same direction. At 15--65$^\circ$, they range from $23^\circ$ to $90^\circ$ (median $47^\circ$): the direction of a single step is then largely set by the run-to-run dispersion. All subsequent results use the 15/65$^\circ$ protocol, which also aligns the IFT cost with the validation test.

Valve~B behaves differently under the same protocols. Its CV is 5.7\% at 30--60$^\circ$ and 4.3\% at 15--65$^\circ$, and at 15--65$^\circ$ all single-shot gradients lie within 10$^\circ$ of their mean $\hat g=[-0.5\pm1.7,\,-13.0\pm1.9]$ in $(\log K,\log\alpha)$. The two valves thus offer two regimes of the same plant: on valve~A, the $K$ component of the gradient is buried in noise, whereas on valve~B both components are consistent from one iteration to the next.

\subsection{Gradient check on valve A}

To check the IFT gradient against a model-free reference, the cost was also measured at $\vartheta\pm h\,\mathbf e_j$, $h=0.3$, for each coordinate $j$ of $\vartheta$, each evaluation being one run of experiment~1; the central differences $(J(\vartheta+h\mathbf e_j)-J(\vartheta-h\mathbf e_j))/(2h)$ estimate the gradient of the measured cost, friction effects included. At the hand-tuned and PI controllers of valve~A, eight interleaved repetitions of these finite-difference experiments and of the IFT experiments ($c\in\{0.5,1,2\}$) were performed (Fig.~\ref{fig:bgrad}, Appendix~\ref{app:bench}). At the hand-tuned controller, the finite differences give $\nabla J=[-4.8\pm1.2,\,-5.7\pm1.8]$ (mean $\pm$ standard error), while IFT with $c=1$ gives $[1.0\pm1.0,\,-5.2\pm1.0]$: the $\alpha$ component is correct, but the $K$ component is missed. The cosine between both remains positive ($0.64$, with a 90\% bootstrap interval of $[0.33,0.87]$), yet a step of 0.3 along $-\hat g/\|\hat g\|$ does not decrease the measured cost. Eight repetitions of $J$ before and after the step differ by $-0.4\%$ only, and a one-sided Welch $t$-test of the hypothesis that the cost decreased gives a $p$-value of 0.46, i.e.\ the decrease is indistinguishable from the run-to-run dispersion: the plateau of Section~\ref{sec:opt} is observed directly. At the PI controller, farther from the optimum, the IFT direction is reliable (cosine 0.84--0.88 for all $c$) and the same step decreases $J$ by 5.1\% ($p=0.07$).

\subsection{Tuning: the best update rule depends on the valve}

\begin{figure}[!t]
\centering
\includegraphics[width=\textwidth]{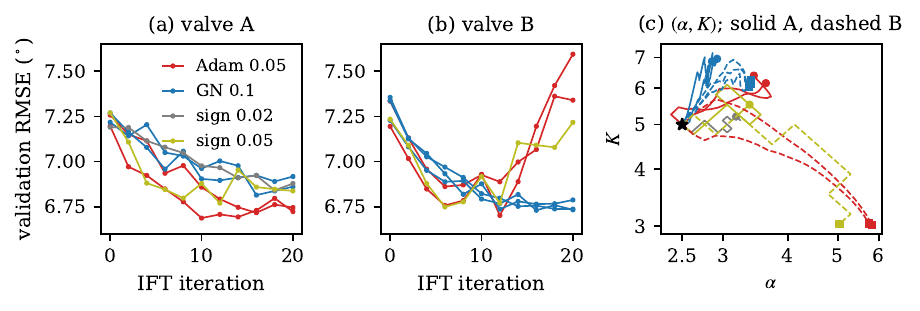}
\caption{Hardware tuning runs (15/65$^\circ$, 20 iterations): validation RMSE every second iteration on valve~A (a) and valve~B (b), and trajectories in the $(\alpha,K)$ plane (c; $\star$: hand-tuned start; markers: end points).}
\label{fig:btune}
\end{figure}

On valve~A, GN ($\gamma=0.1$) and Adam ($\mu=0.05$) were each run twice for 20 iterations, and the sign step once with $h=0.02$ and once with $h=0.05$ (Fig.~\ref{fig:btune}a,c). The measured $J$ (21--31) is useless as a progress indicator, whereas the validation RMSE is reproducible to within $0.02$--$0.08^\circ$ (Table~\ref{tab:bench}, Appendix~\ref{app:bench}) and decreases steadily. Each method converges to its own reproducible region, GN to $(6.9,2.9)$ and Adam to $(6.2$--$6.4,3.4$--$3.6)$, with a validation RMSE over the last three evaluations of 6.84--6.91$^\circ$ for GN and 6.74--6.75$^\circ$ for Adam, from 7.20--7.27$^\circ$ initially.

\begin{figure}[!t]
\centering
\includegraphics[width=\textwidth]{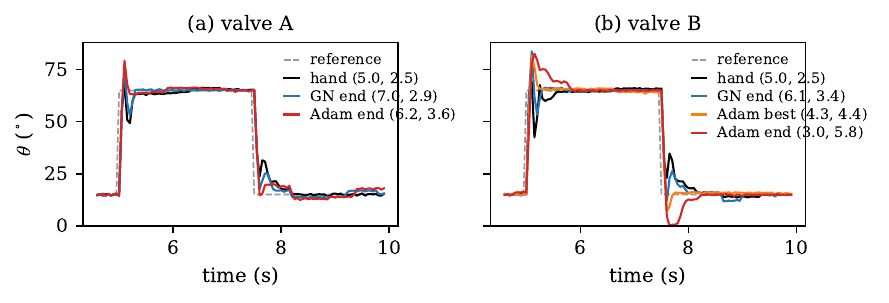}
\caption{Closed-loop responses during the validation test (one up and one down step) for the hand-tuned controller, the end points of GN and Adam, and, on valve~B, the best Adam iterate; $(K,\alpha)$ in parentheses.}
\label{fig:bresp}
\end{figure}

These runs separate the two ingredients of Adam. The $\alpha$ component of $\hat g$ is consistently negative, but $\hat R^{-1}$ shrinks this direction, so that GN moves $\log\alpha$ by only $0.007$ per iteration on average, against $0.016$--$0.019$ for Adam and a steady increase for the sign step, which has no momentum: the normalisation frees the direction that GN under-weights. The $K$ component of $\hat g$, on the other hand, is dominated by noise (Fig.~\ref{fig:bgrad}). GN reverses its $K$ step in 37--53\% of the iterations and travels 2.3--2.6 times the net distance in $\log K$, and the sign step with $h=0.02$ performs a random walk ($K\in[4.8,5.2]$, path ratio 10), whereas Adam extracts the weak drift (reversals 11--32\%, path ratio 1.2--1.4). A larger normalised step does not reproduce this behaviour: with $h=0.05$, the sign step reaches $\alpha=3.37$ within six iterations, as Adam does, but $K$ wanders between 4.8 and 6.1 (42\% reversals) and the final RMSE ($6.85^\circ$) is that of GN. In this low signal-to-noise regime, Adam thus combines a conditioning effect of the normalisation on one coordinate and a variance-reduction effect of the momentum on the other. The improvement of the quadratic criterion comes with more overshoot, however (Fig.~\ref{fig:bresp}a): Adam gives the lowest RMSE but 29\% overshoot on the upward steps, against 20\% and the fastest settling for GN (Table~\ref{tab:bench}). A masked criterion that ignores the transient \citep{lequin2003iterative} would trade these differently.

On valve~B, the ranking is reversed (Fig.~\ref{fig:btune}b,c and Table~\ref{tab:bench}). GN was run three times and converged each time to the same controller, $(6.06,3.36)$, $(6.25,3.40)$ and $(6.15,3.40)$, whose validation RMSE is $6.71^\circ$, against $7.25^\circ$ for the hand-tuned controller. Adam was run twice and drifted both times to $(3.0,5.8)$, whose RMSE, $7.46^\circ$, is worse than that of the hand-tuned controller, although its best iterate along the way, $(4.27,4.42)$, is as good as GN ($6.72^\circ$). With a low integral gain $K/\alpha\approx0.5$, the final Adam controller stays above the set point for about half a second after the upward step, instead of a brief spike for GN, and hits the lower stop after the downward step (Fig.~\ref{fig:bresp}b). The drift is not an artefact of the validation test. Ten repetitions of the IFT experiment at both end points give $J=20.6\pm0.9$ at the GN end point and $21.6\pm1.1$ at the Adam end point (Welch test, $p=0.03$), and the IFT gradient at the Adam end point, $[-1.0\pm1.8,\,+7.8\pm2.1]$, correctly points back towards a larger $K$ and a smaller $\alpha$. Adam has thus overshot a region that the gradient identifies correctly. On this valve, the $K$ component of $\hat g$ is small but consistently positive (in 80--85\% of the iterations, mean $+2$ against about $-11$ for the $\alpha$ component). GN, through the coupling in $\hat R^{-1}$, increases $K$ by about 0.01 per iteration despite the sign of $\hat g_K$, whereas a coordinate-wise normalisation gives this small component the same weight as the $\alpha$ component. The sign step with $h=0.05$ isolates this effect: without momentum, it also decreases $K$ steadily, to 3.0, and its validation RMSE rises from 6.75 to $7.22^\circ$, while $\alpha$ oscillates around 5, where the $\alpha$ component of $\hat g$ changes sign. Adam follows the same path and, because of its momentum, carries $\alpha$ beyond this point, to 5.8. On valve~B, the normalisation is therefore the cause of the drift, and the momentum amplifies it.

\subsection{Twin versus hardware}

The protocol was also run on the twins of both valves, with 0.1$^\circ$ noise and $\pm20\%$ friction variability. The twin of valve~A reproduces the cost level of the 30--60$^\circ$ experiment ($J=17.6$ against 17.0), the ranking of the controllers of Table~\ref{tab:bench}, their overshoot and the cosines of Fig.~\ref{fig:bgrad}a, but not the structure of the run-to-run variability (CV of 6\% and 16\% at 30/60 and 15/65, against 19\% and 6\% on hardware). On this twin, Adam degrades the validation error by decreasing $K$, as on valve~B. The twin of valve~B fails more clearly: at the hand-tuned controller, it predicts a stick--slip limit cycle between 36 and 90$^\circ$ around the 65$^\circ$ set point, which the valve does not exhibit, and hence a validation RMSE of $12.6^\circ$ instead of $7.3^\circ$. It ranks the Adam end point first ($6.7^\circ$) and the GN end point last among the tuned controllers ($7.3^\circ$), the reverse of the hardware. The twins thus expose the mechanism but cannot select the update rule for a given unit.

\section{Discussion and conclusion}

IFT is stochastic gradient descent with an experimental oracle, and on a plant with stiction this oracle is biased. On throttle valves, the bias stems from the initial-state mismatch that stiction imposes on the two experiments of one iteration (Proposition~\ref{prop:bias}); it was quantified on digital twins and observed on hardware. As for any biased stochastic gradient method, progress stalls and continuing may degrade the controller, so that a stopping rule based on a separate validation test, rather than on the noisy IFT cost, is indispensable; on valve~B, it would have stopped Adam close to its best evaluation, at the level reached by GN.

Two lessons seem transferable to other data-driven tuning problems on mechanical plants. The first is that the design of the experiment matters more than the update rule: spanning the operating range of interest and running the second experiment away from the stops reduced the dispersion of the cost from 19 to 6\% on valve~A and made the IFT cost consistent with the validation criterion. The second is that adaptive updates are neither good nor bad in themselves: their coordinate-wise normalisation replaces the GN metric, which helps when that metric discounts a reliable direction and hurts when it discounts an unreliable one, and their momentum pays off only when a weak trend is buried in noise. Two valves of the same reference and the same protocol produced both situations, and their twins did not predict which one would occur; the twin of valve~B even ranked the controllers in reverse order.

The hardware study is limited to two valves and two or three runs per method and valve, and the proposition covers a constant offset only. Natural continuations are a gradient-free baseline such as Bayesian optimisation \citep{berkenkamp2016safe}, friction-aware estimators (dithering, repeated settling, or a correction of \eqref{eq:bias} with an estimate of $d$), and masked criteria \citep{lequin2003iterative} to reduce the overshoot.

\bibliography{references}

\begin{thebibliography}{24}
\providecommand{\natexlab}[1]{#1}
\providecommand{\url}[1]{\texttt{#1}}
\expandafter\ifx\csname urlstyle\endcsname\relax
  \providecommand{\doi}[1]{doi: #1}\else
  \providecommand{\doi}{doi: \begingroup \urlstyle{rm}\Url}\fi

\bibitem[Ajalloeian and Stich(2020)]{ajalloeian2020convergence}
A.~Ajalloeian and S.~U. Stich.
\newblock On the convergence of {SGD} with biased gradients.
\newblock \emph{arXiv preprint arXiv:2008.00051}, 2020.

\bibitem[Baciu and Lazar(2023)]{baciu2023iterative}
A.~Baciu and C.~Lazar.
\newblock Iterative feedback tuning of model-free intelligent {PID}
  controllers.
\newblock \emph{Actuators}, 12\penalty0 (2):\penalty0 56, 2023.
\newblock \doi{10.3390/act12020056}.

\bibitem[Baciu et~al.(2021)Baciu, Lazar, and Caruntu]{baciu2021iterative}
A.~Baciu, C.~Lazar, and C.-F. Caruntu.
\newblock Iterative feedback tuning of model-free controllers.
\newblock In \emph{Proceedings of the 25th International Conference on System
  Theory, Control and Computing (ICSTCC)}, pages 467--472, 2021.

\bibitem[Belhadjoudja et~al.(2023)Belhadjoudja, Maghenem, and
  Witrant]{belhadjoudja2023intelligent}
M.~C. Belhadjoudja, M.~Maghenem, and E.~Witrant.
\newblock On the intelligent proportional controller applied to linear systems.
\newblock In \emph{Proceedings of the 62nd IEEE Conference on Decision and
  Control (CDC)}, pages 938--943, 2023.

\bibitem[Berkenkamp et~al.(2016)Berkenkamp, Schoellig, and
  Krause]{berkenkamp2016safe}
F.~Berkenkamp, A.~P. Schoellig, and A.~Krause.
\newblock Safe controller optimization for quadrotors with {G}aussian
  processes.
\newblock In \emph{Proceedings of the IEEE International Conference on Robotics
  and Automation (ICRA)}, pages 491--496, 2016.

\bibitem[Bernstein et~al.(2018)Bernstein, Wang, Azizzadenesheli, and
  Anandkumar]{bernstein2018signsgd}
J.~Bernstein, Y.-X. Wang, K.~Azizzadenesheli, and A.~Anandkumar.
\newblock {signSGD}: compressed optimisation for non-convex problems.
\newblock In \emph{Proceedings of the 35th International Conference on Machine
  Learning (ICML)}, volume~80 of \emph{Proceedings of Machine Learning
  Research}, pages 560--569, 2018.

\bibitem[Campi et~al.(2002)Campi, Lecchini, and Savaresi]{campi2002virtual}
M.~C. Campi, A.~Lecchini, and S.~M. Savaresi.
\newblock Virtual reference feedback tuning: a direct method for the design of
  feedback controllers.
\newblock \emph{Automatica}, 38\penalty0 (8):\penalty0 1337--1346, 2002.
\newblock \doi{10.1016/S0005-1098(02)00032-8}.

\bibitem[Canudas~de Wit et~al.(2001)Canudas~de Wit, Kolmanovsky, and
  Sun]{Canudas2001}
C.~Canudas~de Wit, I.~Kolmanovsky, and J.~Sun.
\newblock Adaptive pulse control of electronic throttle.
\newblock In \emph{Proceedings of the 2001 American Control Conference},
  volume~4, pages 2872--2877, 2001.
\newblock \doi{10.1109/ACC.2001.946335}.

\bibitem[Fliess and Join(2013)]{fliess2013model}
M.~Fliess and C.~Join.
\newblock Model-free control.
\newblock \emph{International Journal of Control}, 86\penalty0 (12):\penalty0
  2228--2252, 2013.
\newblock \doi{10.1080/00207179.2013.810345}.

\bibitem[Hjalmarsson(1998)]{hjalmarsson1998nonlinear}
H.~Hjalmarsson.
\newblock Control of nonlinear systems using iterative feedback tuning.
\newblock In \emph{Proceedings of the 1998 American Control Conference}, pages
  2083--2087, 1998.

\bibitem[Hjalmarsson(2002)]{hjalmarsson2002overview}
H.~Hjalmarsson.
\newblock Iterative feedback tuning -- an overview.
\newblock \emph{International Journal of Adaptive Control and Signal
  Processing}, 16\penalty0 (5):\penalty0 373--395, 2002.
\newblock \doi{10.1002/acs.714}.

\bibitem[Hjalmarsson et~al.(1998)Hjalmarsson, Gevers, Gunnarsson, and
  Lequin]{hjalmarsson1998iterative}
H.~Hjalmarsson, M.~Gevers, S.~Gunnarsson, and O.~Lequin.
\newblock Iterative feedback tuning: theory and applications.
\newblock \emph{IEEE Control Systems Magazine}, 18\penalty0 (4):\penalty0
  26--41, 1998.
\newblock \doi{10.1109/37.710876}.

\bibitem[Jiao et~al.(2014)Jiao, Zhang, and Shen]{Jiao2014}
X.~Jiao, J.~Zhang, and T.~Shen.
\newblock An adaptive servo control strategy for automotive electronic throttle
  and experimental validation.
\newblock \emph{IEEE Transactions on Industrial Electronics}, 61\penalty0
  (11):\penalty0 6275--6284, 2014.
\newblock \doi{10.1109/TIE.2014.2311398}.

\bibitem[Karimi et~al.(2008)Karimi, Butcher, and Longchamp]{karimi2008model}
A.~Karimi, M.~Butcher, and R.~Longchamp.
\newblock Model-free precompensator tuning based on the correlation approach.
\newblock \emph{IEEE Transactions on Control Systems Technology}, 16\penalty0
  (5):\penalty0 1013--1020, 2008.

\bibitem[Karnopp(1985)]{karnopp1985}
D.~Karnopp.
\newblock Computer simulation of stick-slip friction in mechanical dynamic
  systems.
\newblock \emph{Journal of Dynamic Systems, Measurement, and Control},
  107\penalty0 (1):\penalty0 100--103, 1985.
\newblock \doi{10.1115/1.3140698}.

\bibitem[Kingma and Ba(2015)]{Kingma2014AdamAM}
D.~P. Kingma and J.~Ba.
\newblock Adam: a method for stochastic optimization.
\newblock In \emph{Proceedings of the 3rd International Conference on Learning
  Representations (ICLR)}, 2015.

\bibitem[Lequin et~al.(2003)Lequin, Gevers, Mossberg, Bosmans, and
  Triest]{lequin2003iterative}
O.~Lequin, M.~Gevers, M.~Mossberg, E.~Bosmans, and L.~Triest.
\newblock Iterative feedback tuning of {PID} parameters: comparison with
  classical tuning rules.
\newblock \emph{Control Engineering Practice}, 11\penalty0 (9):\penalty0
  1023--1033, 2003.
\newblock \doi{10.1016/S0967-0661(02)00303-9}.

\bibitem[Mart{\'i}nez-Jim{\'e}nez et~al.(2026)Mart{\'i}nez-Jim{\'e}nez,
  Miller-Murphy, Ondoa-Essomba, and Witrant]{martinez2026code}
F.~Mart{\'i}nez-Jim{\'e}nez, S.~Miller-Murphy, T.~Ondoa-Essomba, and
  E.~Witrant.
\newblock Learning to tune with biased gradients: code, firmware and data for
  iterative feedback tuning of model-free control on throttle valves.
\newblock Zenodo, 2026.
\newblock URL \url{https://doi.org/10.5281/zenodo.23493908}.

\bibitem[Munteanu et~al.(2025)Munteanu, Pintilie, and
  Lazar]{munteanu2025practical}
M.-D. Munteanu, D.-I. Pintilie, and C.~Lazar.
\newblock Practical issues in the implementation of {IFT} algorithm for
  model-free controllers.
\newblock In \emph{Proceedings of the 25th International Conference on Control
  Systems and Computer Science (CSCS)}, pages 80--87, 2025.

\bibitem[Proch{\'a}zka et~al.(2005)Proch{\'a}zka, Gevers, Anderson, and
  Ferrera]{prochazka2005iterative}
H.~Proch{\'a}zka, M.~Gevers, B.~D.~O. Anderson, and C.~Ferrera.
\newblock Iterative feedback tuning for robust controller design and
  optimization.
\newblock In \emph{Proceedings of the 44th IEEE Conference on Decision and
  Control and European Control Conference}, pages 3602--3607, Seville, Spain,
  2005.
\newblock \doi{10.1109/CDC.2005.1582721}.

\bibitem[Pujol et~al.(2016)Pujol, Vidal, Acho, and Vargas]{Pujol2016}
G.~Pujol, Y.~Vidal, L.~Acho, and A.~N. Vargas.
\newblock Asymmetric modelling and control of an electronic throttle.
\newblock \emph{International Journal of Numerical Modelling: Electronic
  Networks, Devices and Fields}, 29\penalty0 (2):\penalty0 192--204, 2016.
\newblock \doi{10.1002/jnm.2063}.

\bibitem[Sj{\"o}berg and Agarwal(1996)]{sjoberg1996model}
J.~Sj{\"o}berg and M.~Agarwal.
\newblock Model-free repetitive control design for nonlinear systems.
\newblock In \emph{Proceedings of the 35th IEEE Conference on Decision and
  Control}, volume~4, pages 2824--2829, Kobe, Japan, 1996.

\bibitem[Witrant et~al.(2023)Witrant, Landau, and Vaillant]{witrant2023data}
E.~Witrant, I.~D. Landau, and M.-P. Vaillant.
\newblock A data-driven control methodology applied to throttle valves.
\newblock \emph{Control Engineering Practice}, 139:\penalty0 105634, 2023.

\bibitem[Yahagi and Kajiwara(2022)]{yahagi2022non}
S.~Yahagi and I.~Kajiwara.
\newblock Non-iterative data-driven tuning of model-free control based on an
  ultra-local model.
\newblock \emph{IEEE Access}, 10:\penalty0 72773--72784, 2022.

\end{thebibliography}

\appendix

\FloatBarrier
\section{Proof of Proposition~\ref{prop:bias}}\label{app:proof}
Write all signals of experiment~2 in deviation from $(\theta_2,u_{2,0})$: $\tilde y=y_2-\theta_2$, $\tilde u=u_2-u_{2,0}$ and $\tilde r=r_2-\theta_2=-c\,e_1$. Since $u_{2,0}$ is the initial value of the integral state and the loop starts at rest with $e=0$, the incremental law \eqref{eq:ip} gives $\tilde u=C(\tilde r-\tilde y)$ with zero initial conditions, while the plant gives $\tilde y=G(\tilde u+d\,H)$. Eliminating $\tilde u$ yields $\tilde y=T\tilde r+GS\,d\,H$ and $\tilde u=CS\tilde r-T\,d\,H$. Applying $\Phi/c$ and using $\psi^y=-\Phi Te_1$ and $\psi^u=-\Phi CSe_1$ gives the expressions of $\hat\psi^y$ and $\hat\psi^u$. Since $\hat g$ is obtained by inserting $\hat\psi^y$ and $\hat\psi^u$ in the right-hand side of \eqref{eq:dJ}, which is linear in them, \eqref{eq:bias} follows. A measurement noise $n$ in experiment~2 adds $\frac1c \Phi S\,n$ and $-\frac1c \Phi CS\,n$ in the same way, which is the classical result of \citet{hjalmarsson1998iterative}; the offset term is a deterministic, low-frequency analogue of this noise term, with the same $1/c$ scaling, but not zero-mean. \hfill$\square$

\FloatBarrier
\section{Digital twins}\label{app:twin}
\begin{figure}[h!]
\centering
\includegraphics[width=\textwidth]{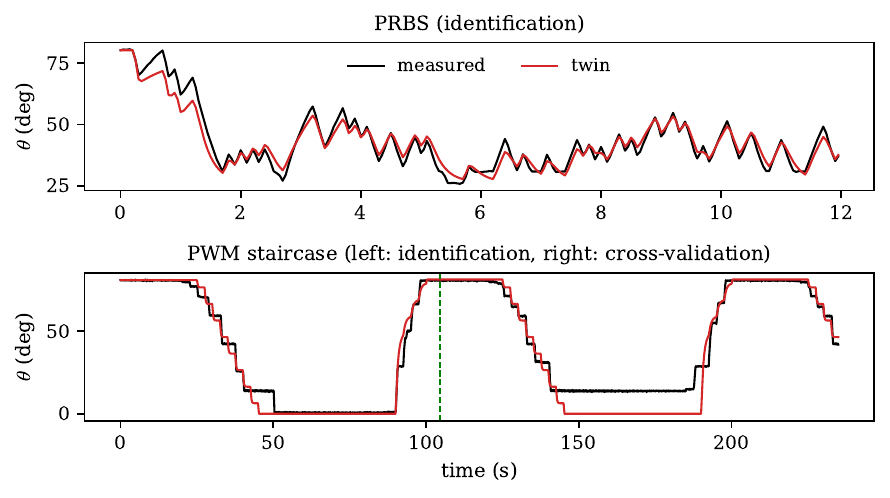}
\caption{Twin of one valve: PRBS used for identification (top) and PWM staircase (bottom; identification on the left of the dashed line, cross-validation on the right).}
\label{fig:twin}
\end{figure}

The model \eqref{eq:twin} is integrated with 50 sub-steps per sampling period and a zero-order hold on $v$. When $\omega=0$, the valve stays at rest as long as the net torque $-a_1(\theta-\theta_s)+a_4v$ lies in $[-a_{3s}^-,a_{3s}^+]$. A reversal of $\omega$ caused by the friction term within a sub-step is replaced by a stop, whereas a reversal caused by the other torques is integrated normally, and the stops are inelastic. The identification criterion is the sum of the RMS simulation errors on the PRBS (511 samples) and on the first staircase cycle, the latter with weight 0.5. One valve has no staircase record and is identified on the PRBS only; it is the outlier of the ablation study. The identified friction asymmetry changes sign between valves, which indicates that the preload and the friction levels are only partially identifiable. Fig.~\ref{fig:twin} shows a typical fit and Table~\ref{tab:twins} gives the parameters of the ten twins; the large spread of the friction parameters, sometimes over two orders of magnitude, reflects both the dispersion of the valves and their partial identifiability. Valves A and B of Section~\ref{sec:bench} are valves 9 and 0.

\begin{table}[h!]
\centering
\scriptsize
\setlength{\tabcolsep}{2.5pt}
\caption{Parameters of the twins (angles in degrees, time in s, input in PWM units) and RMS simulation errors ($^\circ$) on the PRBS, on the staircase cycle used for identification (fit) and on the cross-validation cycles (CV); valve~6 has no staircase record and was identified on the PRBS only.}
\label{tab:twins}
\begin{tabular}{@{}r|rrrrrrrrrrr|rrr@{}}
\toprule
valve & $a_1$ & $\theta_s$ & $a_2^+$ & $a_2^-$ & $a_3^+$ & $a_3^-$ & $a_{3s}^+$ & $a_{3s}^-$ & $a_4$ & $\theta_{\min}$ & $\theta_{\max}$ & PRBS & fit & CV \\
\midrule
0 & 63 & 197 & 52 & 37 & 816 & 13377 & 1154 & 16949 & 364 & 1.3 & 90 & 2.1 & 4.9 & 10.1 \\
1 & 1655 & 177 & 1000 & 357 & 133820 & 87401 & 187340 & 108673 & 4200 & 0.0 & 81 & 2.7 & 6.1 & 6.6 \\
2 & 59 & 107 & 41 & 27 & 619 & 9099 & 1724 & 9297 & 186 & 0.0 & 82 & 2.6 & 6.7 & 13.1 \\
3 & 82 & 101 & 71 & 15 & 1966 & 94 & 5898 & 107 & 73 & 1.6 & 81 & 3.5 & 9.3 & 10.5 \\
5 & 44 & 162 & 23 & 11 & 3266 & 230 & 4245 & 331 & 78 & 13 & 82 & 2.6 & 6.5 & 10.5 \\
6 & 238 & 263 & 1000 & 33 & 15545 & 45951 & 35261 & 46232 & 1284 & 0.0 & 79 & 0.9 & -- & -- \\
7 & 313 & 82 & 190 & 127 & 2784 & 50544 & 4126 & 81420 & 869 & 13 & 79 & 2.1 & 7.9 & 13.2 \\
9 & 69 & 88 & 36 & 20 & 42 & 4482 & 72 & 7371 & 112 & 0.4 & 81 & 2.7 & 6.0 & 13.8 \\
10 & 85 & 168 & 67 & 22 & 6737 & 716 & 10614 & 1006 & 171 & 0.0 & 81 & 2.8 & 4.5 & 9.2 \\
11 & 32 & 278 & 49 & 45 & 3397 & 5263 & 5027 & 6081 & 225 & 0.6 & 90 & 1.7 & 3.3 & 4.9 \\
\bottomrule
\end{tabular}
\end{table}

\FloatBarrier
\section{Additional results on the twins}\label{app:opt}

\begin{table}[h!]
\centering
\small
\caption{Update rules on the ten twins after 40 iterations (80 runs per cell): failure rate, median and 90th percentile of the normalised cost gap ($g=1$: no improvement; $g=0$: optimum).}
\label{tab:opt}
\begin{tabular}{@{}lc|ccc|ccc@{}}
\toprule
 & & \multicolumn{3}{c|}{noise} & \multicolumn{3}{c}{harsh} \\
update & $h$ & fail & $\tilde g_{40}$ & $g^{90}_{40}$ & fail & $\tilde g_{40}$ & $g^{90}_{40}$ \\
\midrule
GN & 0.1 & 0\% & 0.61 & 3.69 & 0\% & 0.81 & 6.90 \\
GN & 0.5 & 36\% & 0.54 & 9.10 & 51\% & 0.82 & 13.14 \\
GN, $\|\Delta\|\le0.2$ & 0.3 & 0\% & 0.85 & 6.81 & 0\% & 0.76 & 11.75 \\
sign step & 0.02 & 0\% & 0.62 & 2.72 & 0\% & 0.62 & 2.72 \\
ADAM & 0.02 & 0\% & 0.50 & 1.76 & 0\% & 0.50 & 2.29 \\
ADAM & 0.05 & 0\% & 0.57 & 3.05 & 0\% & 0.79 & 4.26 \\
ADAM on $\hat R^{-1}\hat g$ & 0.05 & 0\% & 0.53 & 3.66 & 0\% & 0.68 & 4.27 \\
\bottomrule
\end{tabular}
\end{table}

\begin{figure}[h!]
\centering
\includegraphics[width=\textwidth]{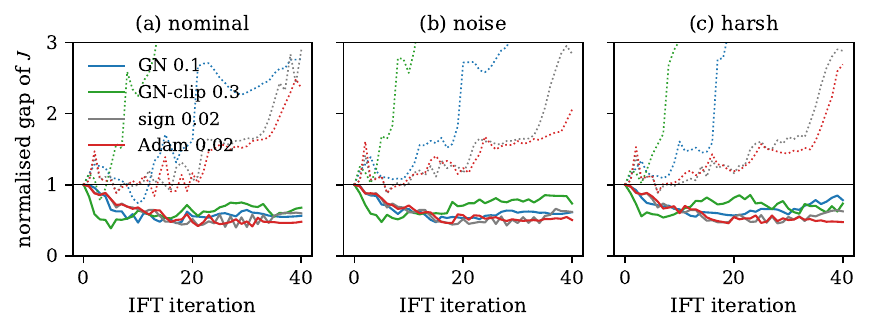}
\caption{Normalised cost gap along the iterations on the twins (solid: median over valves and seeds; dotted: 90th percentile).}
\label{fig:opt}
\end{figure}

The initial controller changes the picture. From an aggressive controller ($K=30$, $\alpha=1.2$), GN reaches a median gap of 0.22 in 25 iterations, and 0.01 with clipping, against 0.88 for Adam with $\mu=0.02$ and 0.04 with $\mu=0.05$. From a sluggish one ($K=0.3$, $\alpha=15$), unclipped GN fails in 38\% of the runs, while clipped GN and Adam with $\mu=0.05$ reach 0.09 and 0.12. The parametrisation matters as well: applied in $(q_0,q_1)$ coordinates, Adam-GN fails within ten iterations for 60\% of the twins with $\mu=1$ and for all of them with $\mu=30$, by cancelling the integral action.

An acceptance test that rejects any step increasing the measured cost leaves a median gap of 0.84--1.00 after 40 iterations, i.e.\ it stops the progress almost entirely.

The statistical comparisons of Section~\ref{sec:opt} are paired Wilcoxon signed-rank tests over valves and seeds on the gap averaged over the last five iterations. Under noise, the median difference between Adam ($\mu=0.02$) and GN ($\gamma=0.1$) is $-0.06$ ($p<10^{-3}$), while the differences with the sign step ($p=0.20$) and with GN at $\gamma=0.05$ ($p=0.99$) are not significant. In the harsh scenario, Adam improves on the sign step by $-0.09$ ($p<10^{-3}$) and on GN at $\gamma=0.05$ by only $-0.01$.

\FloatBarrier
\section{Hardware protocol and additional results}\label{app:bench}
The set-up consists of a throttle valve with a single return spring that opens it, a DC motor driven by a 10\,A H-bridge shield, an Arduino Mega 2560 and a magnetic angle sensor calibrated so that the closed stop reads 0$^\circ$; the PWM saturates at 255. The worst sampling delay observed with the firmware was 12\,$\mu$s, with no overrun.

The firmware accepts a \texttt{RUN} command whose arguments are the type of run (experiment~1, experiment~2 generated on board as $r_2=\theta_2-c\,e_1$, validation, or hold), $K$, $\alpha$, $c$, $\theta_A$, $\theta_B$, $N$, the settling time, $u_{\max}$ and $\theta_2$, and a \texttt{CAL} command for the sensor calibration. Each run starts with 3\,s of closed-loop settling at its initial angle. Experiment~1 follows $\theta_A\to\theta_B\to\theta_A$ over $N=200$ samples (10\,s), with 30/60$^\circ$ in the twin study and 15/65$^\circ$ with $\theta_2=40^\circ$ in the hardware study, and $\lambda=0.01$.

For the gradient check, each controller is evaluated by eight interleaved repetitions of the IFT experiments for $c\in\{0.5,1,2\}$ and of the four finite-difference experiments ($\pm0.3$ in $\log K$ and $\log\alpha$); the cosines are given with 2000-sample bootstrap intervals, and the descent test compares eight repetitions of $J$ at $\vartheta$ and at $\vartheta-0.3\hat g/\|\hat g\|$ with a one-sided Welch test. The tuning runs have 20 iterations with a validation test every second iteration, with $\gamma=0.1$ for GN, $\mu=0.05$ for Adam and $h\in\{0.02,0.05\}$ for the sign step. The path ratio is $\sum_i|\Delta\log K_i|/|\log K_{20}-\log K_0|$, and the reversal rate is the fraction of iterations at which the sign of $\Delta\log K$ changes.

\begin{figure}[h!]
\centering
\includegraphics[width=\textwidth]{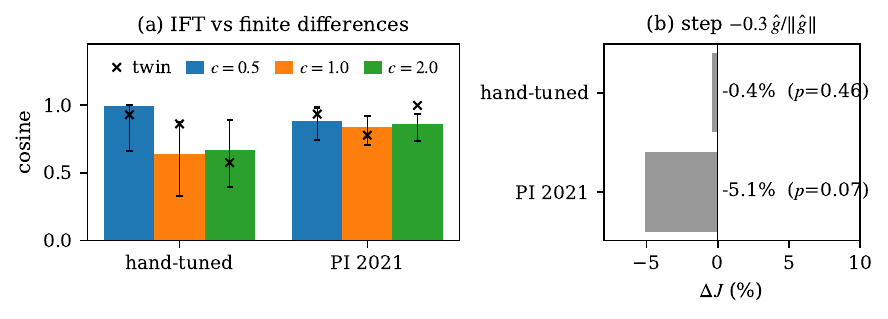}
\caption{Hardware, valve~A, 15/65$^\circ$. (a) Cosine between the IFT gradient and the finite-difference gradient of the measured cost (bars: 90\% bootstrap intervals; crosses: twin of the same valve under the same protocol). (b) Measured change of $J$ after a step of 0.3 along $-\hat g/\|\hat g\|$ ($c=1$) and $p$-value of a one-sided Welch test.}
\label{fig:bgrad}
\end{figure}

Fig.~\ref{fig:bgrad} details the gradient check of Section~\ref{sec:bench} on valve~A, and Table~\ref{tab:bench} gives the results of the validation tests on both valves and on their twins.

\begin{table}[h!]
\centering
\small
\caption{Validation test on hardware (valve~A: 3 repetitions; valve~B: 6 repetitions, end points of two runs pooled): RMSE (mean $\pm$ std), mean overshoot and 2\% settling time of the upward steps, mean steady-state error; last column: twin of the same valve.}
\label{tab:bench}
\begin{tabular}{@{}llcccc|c@{}}
\toprule
 & & RMSE & overshoot & settling & $|e_\infty|$ & twin \\
valve & controller $(K,\alpha)$ & ($^\circ$) & (\%) & (s) & ($^\circ$) & RMSE \\
\midrule
A & hand-tuned (5.00, 2.50) & 7.26$\pm$0.06 & 7 & 1.59 & 0.7 & 7.78 \\
 & GN, end (6.96, 2.92) & 6.82$\pm$0.02 & 20 & 0.64 & 0.4 & 7.65 \\
 & Adam, end (6.16, 3.63) & 6.73$\pm$0.03 & 29 & 1.62 & 1.0 & 7.56 \\
 & PI 2021 (4.41, 5.08) & 7.01$\pm$0.08 & 18 & 2.05 & 1.7 & 7.76 \\
\midrule
B & hand-tuned (5.00, 2.50) & 7.25$\pm$0.03 & 8 & 1.24 & 0.3 & 12.57 \\
 & GN, end (6.15, 3.40) & 6.71$\pm$0.03 & 36 & 0.69 & 0.3 & 7.27 \\
 & Adam, best (4.27, 4.42) & 6.72$\pm$0.06 & 39 & 1.34 & 0.5 & 7.00 \\
 & Adam, end (3.02, 5.78) & 7.46$\pm$0.06 & 26 & 1.82 & 0.4 & 6.68 \\
 & PI 2021 (4.41, 5.08) & 7.14$\pm$0.08 & 40 & 1.70 & 0.5 & 6.73 \\
\bottomrule
\end{tabular}
\end{table}

\FloatBarrier
\end{document}